\documentclass[letterpaper, 10 pt, conference]{ieeeconf}
\usepackage{ifthen}
\IEEEoverridecommandlockouts
\newcommand{\withcoverpage}{1}

\newcommand{\titlestring}{Conformal Prediction of Motion Control
	Performance	for an Automated Vehicle
	in Presence of Actuator Degradations and Failures}
\newcommand{\authorstring}{Richard Schubert,
	Marvin Loba, Jasper Sünnemann, Torben Stolte, and Markus Maurer}
\newcommand{\authorstringaff}{Richard Schubert$^{1}$,
	Marvin Loba$^{1}$, Jasper Sünnemann$^{1}$, Torben Stolte$^{1}$, and Markus Maurer$^{1}$}
\newcommand{\yearstring}{2024}

\title{\LARGE
	\bf\titlestring
	$^{*}$%
	\vspace{-0.5em}
}

\author{\authorstringaff%
	\thanks{$^{*}$\scriptsize%
		This research is accomplished within the
		project
		\autotech\ (FKZ~01IS22088R).
		We acknowledge the financial
		support by %
		the Federal Ministry of Education
		and Research of Germany
		(BMBF).}%
	\thanks{$^{1}$\scriptsize%
		Institute for Control Engineering, TU
		Braunschweig, Germany
			{\tt \{richard.schubert, marvin.loba,
				j.suennemann, t.stolte,
				markus.maurer\}@tu-braunschweig.de}}%
}

\usepackage{verbatim}
\usepackage{tabularx,calc}
\usepackage{color}
\usepackage{tubscolors}
\usepackage{adjustbox}
\usepackage{hyperref}
\usepackage{ifthen}
\usepackage{lipsum}
\usepackage{xstring}
\usepackage{amssymb} 	%
\usepackage{amsmath}
\usepackage{siunitx}
\usepackage{graphicx}
\usepackage{bm} 		%
\usepackage{booktabs} 	%
\usepackage{multirow}   %
\usepackage{booktabs} 	%
\usepackage{colortbl}
\usepackage{microtype}
\usepackage{here}
\usepackage{tabularx}
\usepackage{arydshln}
\usepackage{titlesec}
\usepackage{array}
\usepackage{siunitx}
\usepackage[bibstyle=ieee,
			citestyle=numeric-comp,
			backend=bibtex,
			minnames=1,
            maxnames=3,
			maxcitenames=1,
			maxbibnames=4,
			doi=false,
			isbn=false,
			url=false,
			natbib=true]
			{biblatex}

\DefineBibliographyStrings{english}{andothers={et\addabbrvspace al\adddot}}

\DefineBibliographyStrings{english}{phdthesis = {PhD Th.,}}

\AtEveryBibitem{\clearlist{language}}

\newcommand{\qemph}[1]{\emph{#1}}

\newcommand{\unicar}{UNICAR\emph{agil}}
\newcommand{\autotech}{AUTOtech.\emph{agil}}
\newcommand{\pytorch}{\emph{PyTorch}}

\newcommand{\MATLABSIMULINK}{\emph{MATLAB}}

\newcommand{\mobile}{\emph{MOBILE}}

\newcommand{\ipgc}{\emph{IPG CarMaker}}

\renewcommand{\vec}[1]{\mathbf{#1}} %

\newcommand{\m}{\,\mathrm{m}}
\newcommand{\tm}{\,\mathrm{m}}
\newcommand{\cm}{\,\mathrm{cm}}
\newcommand{\tcm}{\cm}

\newcommand{\toom}{\mathrm{m}^{-1}} %

\newcommand{\tmss}{\,\mathrm{m} / \mathrm{s}^2}

\newcommand{\tkmh}{\,\mathrm{km} / \mathrm{h}}

\renewcommand{\deg}{\,^{\circ}}
\newcommand{\degs}{\,^{\circ}/\mathrm{s}}
\newcommand{\Nm}{\,\mathrm{Nm}}

\newcommand{\odd}{q}

\newcommand{\refe}{^{*}}
\newcommand{\wheel}{w}%
\newcommand{\degradation}{\mathrm{D}}

\newcommand{\zdegwheel}{\tilde{z}_{\degradation, \wheel}}
\newcommand{\deltadegwheel}{\tilde{\delta}_{\degradation, \wheel}}
\newcommand{\deltadotdegwheel}{\dot{\tilde{\delta}}_{\degradation, \wheel}}
\newcommand{\taudegwheel}{\tilde{\tau}_{\degradation, \wheel}}
\newcommand{\nsegments}{N_{\mathrm{S}}}
\newcommand{\nmaneuver}{N_{\mathrm{M}}}
\newcommand{\ndeg}{N_{\degradation}}

\newcommand{\kmin}{k_{\odd}^{\mathrm{min}}}
\newcommand{\kmax}{k_{\odd}^{\mathrm{max}}}
\newcommand{\wmin}{w_{\odd}^{\mathrm{min}}}
\newcommand{\wmax}{w_{\odd}^{\mathrm{max}}}
\newcommand{\amax}{a_{\mathrm{max}}}
\newcommand{\vinitial}{v_{\mathrm{init}}}
\newcommand{\vtarget}{v_{\mathrm{end}}}

\newcommand{\wveh}{w_{\mathrm{Veh}}}

\newcommand{\inputnn}{\xi} %
\newcommand{\inputnntest}{\inputnn_{\mathrm{Test}}}
\newcommand{\inputnnset}{\mathcal{X}}
\newcommand{\inputnntestset}{\inputnnset_{\mathrm{Test}}}
\newcommand{\inputnntrainset}{\inputnnset_{\mathrm{Train}}}
\newcommand{\inputnncalset}{\inputnnset_{\mathrm{Cal}}}

\newcommand{\outputnn}{y}
\newcommand{\outputnntest}{\outputnn_{\mathrm{Test}}}

\newcommand{\interval}{\mathcal{I}}
\newcommand{\suup}{\mathrm{Hi}}
\newcommand{\sulow}{\mathrm{Lo}}
\newcommand{\epsNmax}{\varepsilon_{\mathrm{lat}, \mathrm{max}}}

\newcommand{\epsNmaxHi}{\hat{\varepsilon}_{\mathrm{lat}, \suup}}

\newcommand{\prob}{P}
\newcommand{\coverage}{C_{\alpha}}
\newcommand{\coverageone}{C_{\alpha, 1}}
\newcommand{\coveragetwo}{C_{\alpha, 2}}

\newcommand{\empcoverage}{\hat{C}_{\alpha}}

\newcommand{\epsNmaxFree}{\varepsilon_{\mathrm{lat}, \mathrm{free}}}
\newcommand{\quantile}{\mathrm{Q}}
\newcommand{\mean}{\mathrm{Mean}}
\newcommand{\median}{\mathrm{Median}}

\newcommand{\smallerfootnote}{\fontsize{7}{8.5}\selectfont}
\newcolumntype{C}[1]{>{\centering\arraybackslash}p{#1cm}}
\renewcommand{\normalsize}{\fontsize{10}{11.95}\selectfont} %

\newcommand{\titlebaselineskip}{0.25\baselineskip}
\titlespacing*{\section}{-0.75em}{\dimexpr\titlebaselineskip}{\titlebaselineskip}

\newcommand{\eqskipInPt}{5pt}
\AtBeginDocument{%
  \abovedisplayskip=\eqskipInPt
  \abovedisplayshortskip=\eqskipInPt
  \belowdisplayskip=\eqskipInPt
  \belowdisplayshortskip=\eqskipInPt
}

\titleformat{\paragraph}[runin]{\normalfont\normalsize\itshape}{\theparagraph}{0.5em}{\hspace{0.5em}}[:]
\titlespacing*{\paragraph}{0pt}{1ex plus 0.5ex minus 0.2ex}{0.5em}
\renewcommand{\theparagraph}{\alph{paragraph})}

\ifthenelse{\withcoverpage=1}{
	\pdfoutput=1
\usepackage{hologo}
\usepackage{listings}
\lstset{breaklines,basicstyle=\small,columns=fullflexible,basicstyle=\ttfamily,language={[plain]TeX}}
}{}

\begin{document}

\ifthenelse{\withcoverpage=1}{
\newcommand{\conferencestring}{27th IEEE International Conference on Intelligent Transportation Systems}
\newcommand{\addressstring}{Edmonton, Canada}

\newif\ifpublished
\publishedtrue

\ifpublished
    \twocolumn[
    \begin{@twocolumnfalse}

        \Huge {IEEE Copyright Notice}

        \vspace{0.25cm}
        
        \large {\copyright\ \yearstring\ IEEE. Personal use of this material is permitted. Permission from IEEE must be obtained for all other uses, in any current or future media, including reprinting/republishing this material for advertising or promotional purposes, creating new collective works, for resale or redistribution to servers or lists, or reuse of any copyrighted component of this work in other works.}

        \vspace{1.25cm}
        
        {\Large Published in \emph{\conferencestring}}

        \vspace{0.5cm}

        \textit{Cite as:}
        \vspace{0.2cm}

        \authorstring,
        ``\titlestring,''
        in \emph{\conferencestring},
        \addressstring, \yearstring.

    \end{@twocolumnfalse}
]
\else
    \twocolumn[
        \begin{@twocolumnfalse}
            \textit{Cite as:}
            \vspace{0.2cm}

            \authorstring,
            ``\titlestring,''
            {submitted for publication}.

            \vspace{0.5cm}

            \textit{BibTeX:} \vspace{0.2cm}
        
            \texttt{%
            @inproceedings\{schubert\_conformal\_\yearstring,\\
            author=\{\authorstring\},\\
            title=\{\titlestring\},\\
            year=\{\yearstring\},\\
            publisher=\{submitted for publication\}\\
            \}
            }

        \end{@twocolumnfalse}
    ]
\fi
}{}

\maketitle
\thispagestyle{empty}
\pagestyle{empty}

\vspace{-1.5em}

\begin{abstract}
	Automated driving systems require monitoring mechanisms to ensure safe operation,
especially if system components degrade or fail. %
Their runtime self-representation plays a key role %
as it provides \mbox{a-priori} knowledge about the system's capabilities and limitations.
In this paper, we propose a data-driven approach for deriving such
a self-representation model for the motion controller of an automated vehicle.
A conformalized prediction model is learned and allows estimating how operational conditions as well as
potential degradations and failures of the vehicle's actuators impact motion control performance. %
During runtime behavior generation, our predictor provides a heuristic for determining the admissible action space.
\end{abstract}

\begin{figure}[H]
    \centering
    \includegraphics[width=0.9\linewidth]{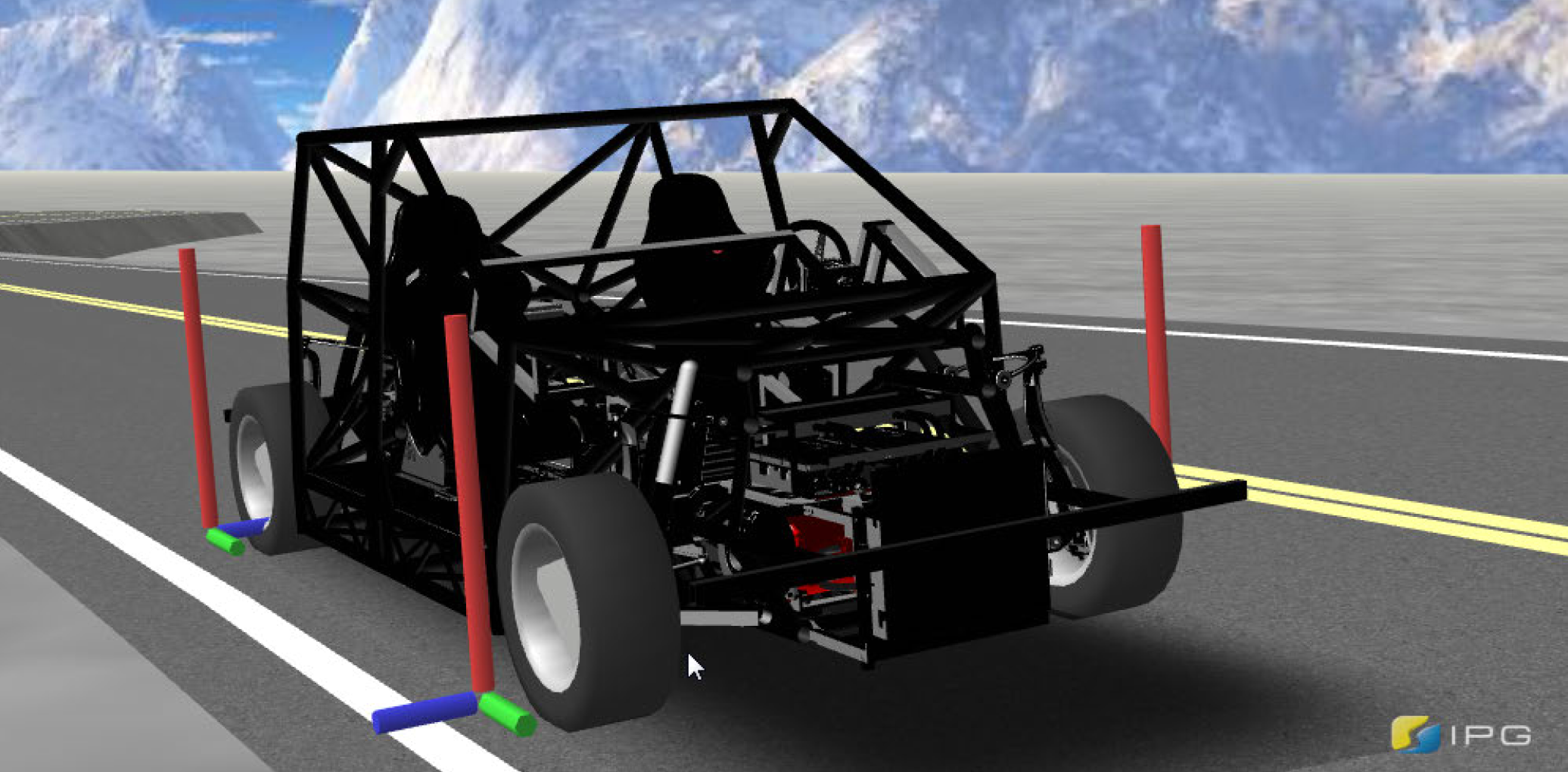}
    \caption{The \mobile\ research vehicle in \ipgc, figure from \parencite{stolte_toward_2023}.}
    \label{fig:01_introduction:01_introduction}
\end{figure}
\vspace{-1.00em}
\section{Introduction} \label{sec:intro}
The development of automated driving technology has the potential of (temporarily)
relieving drivers of the task of driving and monitoring their vehicle, even in challenging scenarios.
However, in order to ensure that these systems operate as desired,
it is necessary to consider safety-related constraints, both during design and operation.
One key consideration when designing an automated driving system is ensuring
that the vehicle can continue to operate safely -- among others -- if hardware failures arise.
Hence, functional boundaries must be carefully evaluated at design
time regarding both normal operation and in the event of a degradation or failure.
In the sense of the ISO~26262 standard~\parencite{international_organization_for_standardization_iso_2018}, both failure
and degradation are the result of a fault, though with different outcome. 
A \qemph{degradation} is the state of a system element with reduced performance, yet with given functionality%
\footnote{Please note, that the understanding of \qemph{degradation} in ISO~26262 is inconclusive, as neither functionality nor performance are defined in the standard.
	Hence, we use the understanding developed in a previous work~\parencite{stolte_taxonomy_2021}.}.
In contrast, a \qemph{failure} denotes the termination of the system's intended behavior, i.e., the system is not providing its functionality.
In addition to robust design, it is necessary to monitor the system's performance at runtime
in order to detect, isolate, and compensate for failures~\parencite{nolte_supporting_2020}
before they lead to hazardous situations.

Furthermore, to adapt behavioral decisions at runtime accordingly,
the system relies on the explicit representation of knowledge about itself,
which is referred to as \qemph{self-representation} \parencite{nolte_supporting_2020}.
This involves creating and deploying models that contain \mbox{a-priori} knowledge
about the system's operational limits.
The impact thereof lies in the ability to, e.g.,
trace the impact of degraded or failed components on the overall system behavior.
By estimating functional boundaries with respect to performance constraints %
of the overall system,
it is possible to determine the admissible actions. %

In this paper, we acknowledge the importance of self-representation
and provide an example of how to derive a self-representation model
of an automated vehicle's motion controller in a data-driven manner.
Given previous work from our group \parencite{stolte_toward_2023},
we use a complex fault-tolerant model-predictive trajectory tracking controller
that is able to control a highly over-actuated vehicle as a case study.
Modelling the dynamics of the controlled vehicle under various
actuator degradations and failures is complex as simple physical models %
might not be sufficient to capture the system's behavior
-- especially when considering over-actuated vehicles.
Based on the results of a large-scale numerical simulation,
we are able to derive a prediction model that estimates
how operational conditions, dynamic constraints,
and potential degradations and failures of the vehicle's actuators impact
the controller's performance at runtime in terms of its deviation from a planned trajectory.
In an example scenario, we show how the prediction model can be used
to estimate the controller's performance in order to
evaluate the feasibility of actions under given conditions, including evasive maneuvers.
This estimation can then be utilized at the behavior generation level
in order to constrain the admissible action space.

The remainder of this paper is structured as follows:
The following \autoref{sec:literature} provides an overview of related work and our contributions.
In \autoref{sec:controller}, we describe the motion controller's role in the functional architecture
and present the fault-tolerant controller by \parencite{stolte_toward_2023}.
Thereafter, we describe our simulation framework and design a predictor
using the generated data in \autoref{sec:predictor}.
Finally, we illustrate the application of the predictor in \autoref{sec:application}.
\section{Related Work} \label{sec:literature}
Previous publications discuss the need to supervise
the behavior generation and execution of automated vehicles:
\textcite{maurer_flexible_2000} argues that any system component should provide quality measures that
must be aggregated by a performance monitoring function and assessed when behavioral decisions are made.
\textcite{nolte_supporting_2020} point out that a system requires a self-representation
in order to determine its capabilities \parencite{reschka_fertigkeiten-_2017}
and select actions to respond appropriately to the current situation.
The set of appropriate actions can be referred to as \textit{admissible action space}
\parencite{nolte_supporting_2020}.
In the context of automated driving, we equate the terms \textit{action} and \textit{driving maneuver},
i.e., an abstraction of possible state progressions for the vehicle's motion \parencite{jatzkowski_zum_2021}.
The constraints of the action space can be both \textit{internal} and \textit{external} in nature,
i.e., related to the system's internal state and the current situation \parencite{nolte_supporting_2020}.

In \parencite{nolte_towards_2017}, \citeauthor{nolte_towards_2017} propose to apply the concept of \qemph{skill graphs}
as introduced by \textcite{reschka_fertigkeiten-_2017} to identify the capabilities of an automated vehicle
and refine an ISO~26262 compliant design process.
This allows to structure system capabilities, assign functional components that contribute to them,
and derive safety requirements. %
Comparable work is presented in the context of the \unicar\ project by \textcite{stolte_towards_2020}.
Both publications propose eliciting requirements related to the vehicle's behavior during system design.
These requirements are then refined when considering the vehicle's architecture.
Behavioral requirements address quantities related to
the vehicle's (externally observable) motion, e.g., the required maximum lateral
deviation from a reference path \parencite{nolte_towards_2017}.
Technical determinants -- e.g., the accuracy of the vehicle's state estimation
or the available performance of the vehicle's actuators --
influence the quality of the vehicle's behavior execution.
In particular, \textcite{nolte_towards_2017} consider an automated vehicle's objective %
to follow a specific lane marking on a German highway:
The authors require that a certain lateral deviation threshold
is not exceeded by the vehicle (with respect to the lane marking).
They use a capability graph to successively refine and decompose into further requirements.
Focusing on capabilities required for behavior execution,
dependencies with respect to the vehicle's localization and motion control capabilities are identified.
Inaccurate localization and/or insufficient performance of motion control
impact behavior execution and may lead to a violation of requirements.
Hence, when considering motion control as an example,
it is crucial to ensure that the controller's performance stays within defined bounds.
Note that the consideration of other sources of uncertainty, e.g., measurement noise,
is not explicitly addressed in this work but should be considered in future research.

In this paper, a particular focus is set on highly over-actuated vehicles and their motion control.
In this context, \textcite{da_silva_crash-prone_2024} investigate the impact of actuator fault combinations
on an over-actuated vehicle's ability to perform an evasive lane change maneuver.
Their offline analysis employs a simulation of the controlled vehicle to acquire several performance metrics,
including the maximum lateral deviation from the reference as well as ``leave road'' and ``collision'' indicators.
In a statistical analysis, they trace the impact of actuator degradations and failures on these metrics
-- however, only for offline analysis.

At runtime, the vehicle should be able to adapt its planned actions and respect its current control capabilities.
Degradations and failures thereof may lead to sudden inconsistencies between the planned and actual motion.
\textcite{ratner_operating_2023} further elaborate on the challenge
of ensuring consistency between (state-based) motion planning and control models for robots.
They propose to bias the planner away from inaccuracies
by increasing the cost of states and actions where discrepancies are observed.
The authors use a stochastic discrepancy model that is learned online from real-world feedback.
A similar approach is presented in \parencite{noseworthy_active_2021}.
In safety-critical applications, however, using online learning to fit such a model
is usually not feasible since model discrepancies must be experienced first before they can be learned.
Therefore, \textcite{castaneda_recursively_2023} pair a Gaussian Process model
with a Control Barrier Function to ensure that the system stays within defined bounds under uncertainty.
Moreover, they propose to acquire new measurements to update their model
only in situations where the underlying dynamics are expected
to only slightly deviate from the model.
To evaluate the space of all actions that can be executed by the system given its actual dynamics,
\textcite{mcconachie_learning_2020} propose the design of a neural network classifier that
predicts the likelihood of a system's action being \qemph{reliable} or \qemph{unreliable}.
An unreliable action is one where the system is likely to run into constraint boundaries
in the real-world that are not considered in a simplified model used for planning.
\citeauthor{mcconachie_learning_2020} exploit this concept to be able
to plan actions in a simplified state space and sort out unreliable actions before executing them.

\begin{figure}
    \centering
    \vspace{0.5em}
    \includegraphics[width=0.9\columnwidth]{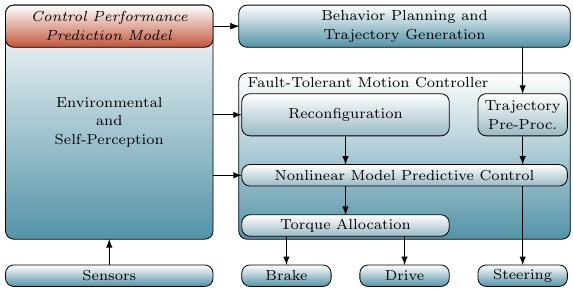}
    \vspace{-0.8em}
    \caption{Simplified functional architecture of an automated vehicle,
        figure taken from \textcite{stolte_toward_2023},
        originally based on \parencite{ulbrich_towards_2017}.
        Our conformalized control performance prediction model relies on self-perception
        information and is located next to the behavior generation block.%
        }
    \label{fig:funcarch}
\end{figure}
Ensuring that a controlled system stays within defined bounds
is also the objective of \qemph{reachability analysis}.
Reachability analysis can be used as a design tool offline
as well as for \qemph{online verification} of dynamic systems
\parencite{althoff_reachability_2010, althoff_online_2014}.
Given model assumptions about the controlled system,
reachability analysis is used to explore which states can be reached (or avoided) for given inputs and disturbances
within a certain time frame or at a specific time point. %
\qemph{Forward} reachable states can be found by propagating the set
of inputs through a dynamics model analytically -- e.g., using set- or distribution-based representations
\parencite{althoff_reachability_2010} -- or numerically.
\qemph{Backward} reachability analysis, on the other hand,
aims to identify the set of states from which the
system cannot avoid reaching an unsafe state at a later time point \parencite{leung_infusing_2020},
regardless of whether a control algorithm tries to prevent this.
Recent approaches often adopt \qemph{Hamilton-Jacobi} reachability analysis
\parencite{bansal_hamilton-jacobi_2017}.
However, it is computationally expensive and scales exponentially with the number of states
of the underlying system \parencite{mitchell_overapproximating_2003, bansal_hamilton-jacobi_2017}.
Systems with more than five states are often considered intractable,
especially for online verification \parencite{leung_infusing_2020}.
To reduce computational effort, for instance, linearized models with small numbers of states are used \parencite{althoff_online_2014}.
If disturbances are considered, they are often assumed to be bounded -- either by fixed boundaries or confidence intervals
-- but methods for \qemph{stochastic reachability analysis} exist as well
\parencite{althoff_reachability_2010, vinod_stochastic_2021, shetty_predicting_2021}.
Recent approaches for reducing the computational effort at runtime
include the learning-based online classification of reachable states
\parencite{rubies-royo_classification-based_2019},
their efficient sampling-based approximation \parencite{lew_sampling-based_2021}
as well as decomposition in less complex subsystems \parencite{chen_decomposition_2018}.

While reachability analysis methods often aim to \qemph{guarantee} that certain state regions are reached/avoided,
the results are always based on assumptions about the dynamics model, input space, and additional disturbances.
Therefore, any model-based guarantee should be considered with care since model-based and additional assumptions may not hold in the real world.
For instance, given sudden degradations of the systems, the dynamics representation must be updated accordingly.
This requires additional research on the derivation of dynamics models under the influence
of degradations and failures, e.g., as presented in \parencite{lombaerts_fault_2011, asadi_damaged_2014} for aviation systems.

\paragraph*{Contributions}
In this paper, we investigate the dynamics of a highly over-actuated, automated vehicle --
which can be described as a nonlinear system with up to 10 states (e.g., \parencite{nolte_sensitivity_2020})
-- under the influence of actuator degradations and failures.
Inspired by~\parencite{mcconachie_learning_2020},
we propose to use a lightweight prediction model that is able to
predict whether an automated vehicle can be controlled with sufficiently
low lateral deviation from the planned motion under nominal, degraded, and failed actuator conditions.
Complementary to the online learning approaches in
\parencite{noseworthy_active_2021, ratner_operating_2023, castaneda_recursively_2023},
our approach exploits prior knowledge about the system's dynamics
under nominal, degraded, and failed conditions instead of online learning.
Combining both approaches in the future could be useful in safety-critical applications.
In particular, we use a neural network and capture the complexity and uncertainty
of the prediction by predicting \qemph{regression quantiles} instead of deterministic values.
To ensure statistical rigor, we apply \qemph{conformalized quantile regression} \parencite{romano_conformalized_2019},
which does not require distributional or boundary assumptions. %
Our data-driven approach is simple to conduct and uses
simulation data acquired from a commercial-grade vehicle dynamics simulation model.
\section{Simulation of the Motion Controller} \label{sec:controller}
In this work, we consider the performance of a motion controller under various dynamic conditions
and under the influence of degradations and failures of the vehicle's actuators.
As a case study, we examine the fault-tolerant motion controller by \parencite{stolte_toward_2023}:
\citeauthor{stolte_toward_2023} point out the advantages of fault-tolerant control
as a strategy to widen the motion controller's
functional boundaries with respect to the possibility of actuator degradations and failures.
The nonlinear model-predictive controller is particularly designed for
highly over-actuated vehicles, i.e., with four drivable, brakable and steerable wheels.
\citeauthor{stolte_toward_2023} deploy and assess their controller
in a simulation of \mobile , which is an overactuated electric research vehicle \parencite{bergmiller_towards_2014}.
Over-actuation is, for instance, also investigated in the
\autotech\ \parencite{van_kempen_autotechagil_2023} and \unicar\, projects \parencite{woopen_unicaragil_2018}:
Each of the four \unicar\, vehicles %
yields the same actuator topology as \mobile .

In the functional architecture of an automated vehicle according to
\textcite{ulbrich_towards_2017} (as shown in \autoref{fig:funcarch}),
the motion controller is located below the behavior generation block
and the subsequent generation of a target trajectory.
To inform the behavior generation process, the vehicle relies on scene information
derived by its environmental perception functions --
for instance, incorporating an estimation of the road/lane geometry.
Also, the scene contains information about the ego-vehicle's internal state.
The internal state may contain the dynamic state as well as degradation and failure states
that are identified through failure detection and isolation functionalities (FDI). 
While tracking the target trajectory, 
the controller receives feedback about the vehicle's current state
through the vehicle's state estimation functions,
e.g., its current position, orientation, and speed. %
However, the measured information in every time step is subject to uncertainty,
e.g., stochastic noise and/or a systematic offset \parencite{werling_neues_2011},
leading to a reduced accuracy. %

At runtime, assessing the controller's performance is important to ensure
that safety requirements in terms of behavior execution are met.
We consider the vehicle's deviation from a planned motion
for which the deviation introduced by the controller is a crucial determinant.
However, simply monitoring a deviation metric itself
means that a deviation is only measured if the vehicle is already deviating from its planned motion --
potentially without being able to recover. %
Therefore, using predictive approaches is crucial. %
Hereafter, we present a data-based prediction model designed for this purpose.
This prediction model can act as a heuristic during the behavior generation phase
and is therefore located next to the behavior generation block in \autoref{fig:funcarch}.
In order to train the prediction model, we present our simulation framework hereafter:
Using an offline reference trajectory generator, we generate a large dataset of trajectories
which meet domain-specific and dynamic constraints.
In a commercial-grade simulation, the sampled trajectories can be used as the input for
the controller and the controller's performance is assessed with respect to a deviation metric.
\subsection{Reference Trajectory Generation} \label{sec:refgen}
\begin{table}%
    \centering
    \caption{Simulation/Prediction Inputs and Outputs}
    \vspace{0.35em}
    \begin{adjustbox}{center, keepaspectratio, width=0.9\columnwidth}
    \renewcommand{\arraystretch}{1.25}
    \begin{tabular}{ccccc}
        \toprule
        \multicolumn{2}{c}{\textbf{Input Variables}} & \textbf{Description} & \textbf{Unit} & \textbf{Value Range} \\
        \midrule
        \multirow{9}{*}{\rotatebox[origin=c]{90}{{\smallerfootnote {ODD and Maneuver Parameters}}}} & $r_{\odd}$ & direction & $[1]$ & $\{\mathrm{left},\, \mathrm{right}\}$ \\
        & $\wmin$ & min. lane width & $[\tm]$ & $[2.78,\, 3.44]$ \\
        & $\wmax$ & max. lane width & $[\tm]$ & $[2.89,\, 3.84]$ \\
        & $\kmin$ & min. curvature & $[\toom]$ & $[-5.4\cdot 10^{-4},\, 0]$ \\
        & $\kmax$ & max. curvature & $[\toom]$ & $[0,\, 1.9\cdot 10^{-4}]$ \\
        & $\mu_{\odd}$ & friction indicator & $[1]$ & $[1]$ \\
        & $\vinitial$ & initial speed & $[\tkmh]$ & $[30,\, 50]$ \\
        & $\vtarget$ & final speed & $[\tkmh]$ & $[30,\, 50]$ \\
        & $\amax$ & max. acceleration & $[\tmss]$ & $[1,\, 5]$ \\
        \addlinespace
        \hdashline
        \addlinespace
        \multirow{3}{*}{\rotatebox[origin=c]{90}{\parbox{1.25cm}{\centering\smallerfootnote {Degradations,\\Failures}}}} 
        	& $\deltadegwheel$ 		& steering angle factor & $[1]$ & $[0,\, 1]$ \\
        	& $\deltadotdegwheel$ 	& steering rate factor & $[1]$ & $[0,\, 1]$ \\
        	& $\taudegwheel$ 		& wheel torque factor& $[1]$ & $[0,\, 1]$ \\
        \addlinespace
        \midrule
        \textbf{Output} & $\epsNmax$ & max. lateral deviation & $\tm$ & $[0.015,\, 1.79]$ \\
        \bottomrule
    \end{tabular}
    \end{adjustbox}
    \label{tab:mlp_inputs_outputs}
\end{table}
To generate meaningful input data for the assessment of the motion controller,
we aim to mimic the interface between online behavior planning and control in a numerical offline simulation.
We use the same trajectory generator as in \parencite{schubert_odd-centric_2023}
which is used to generate reference trajectories for a \emph{lane change} maneuver.
The generator utilizes a simple point mass model and respects derived \emph{maneuver parameters} as its mathematical constraints.
Given these constraints, the optimization algorithm generates a time-minimal trajectory for connecting the current and adjacent lane.
We refer to \parencite{schubert_odd-centric_2023} for a detailed description.
All inputs/constraints of the generation are listed in \autoref{tab:mlp_inputs_outputs}.
The generator outputs a sequence of optimal states
\begin{equation}
    \vec{x}{\refe}(t) = \left(s{\refe}(t), v{\refe}(t), a{\refe}_x(t), a{\refe}_y(t), \psi{\refe}(t), \dot{\psi}{\refe}(t)\right)^T %
\end{equation}
that the trajectory tracking controller shall follow,
with time $t$, Frenet arc length $s$, speed $v$, longitudinal acceleration $a_x$, lateral acceleration $a_y$,
yaw angle $\psi$, and yaw rate $\dot{\psi}$.
Superscript $*$ denotes reference values.
The target and end pose of the maneuver are assumed to be located on the current and adjacent lane's center.
Other trajectory generation models could be used here, e.g., explicitly specifying a target pose
instead of minimizing time given dynamic limitations \parencite{werling_optimal_2010}.
In the remainder of this paper, we only focus on the example of a \qemph{lane change}.
Note that the optimization-based generation is not limited to this. %
With respect to our previous work \parencite{schubert_odd-centric_2023},
we extend the trajectory generation approach as follows:
To increase realism, we use a dataset of real road segment geometries,
collected in the inner city ring road of \emph{Braunschweig, Germany}.
Conducting simulations based on these road segments allows us
to respect realistic distributions of road widths and curvatures
within the operational design domain, i.e., Braunschweig's ring road in this example.
For each road segment with two lanes, we receive the lane center lines for the current and adjacent lines.
We assume the distance between the two lane center lines to be equal to the width of each lane.
While the lane's curvature and width are assumed to be always constant in \parencite{schubert_odd-centric_2023},
here, we store the current road segment's minimum and maximum curvature and width $\kmin, \kmax, \wmin, \wmax$
for a more detailed representation.
Given a minimum segment length of $45\tm$, we receive a set of $403$ road segments from the dataset.
The value ranges of the considered indicators are given in \autoref{tab:mlp_inputs_outputs}.

\subsection{Simulation Setup} \label{sec:setup}
We apply each generated reference trajectory to the trajectory tracking controller by \parencite{stolte_toward_2023}
that is implemented in \MATLABSIMULINK\ and uses a \emph{nonlinear model predictive control} (NMPC) approach.
The controller relies on a nonlinear double-track model paired
with an adapted version of the \emph{Fiala} tire model \parencite{hindiyeh_controller_2014}.
The MPC uses control inputs $\vec{u}_{\mathrm{MPC}} = (\tau_{\wheel}, \dot{\delta}_{\wheel})$ to control the states
$\vec{x}_{\mathrm{MPC}} = (s, d, \psi, v_x^\mathrm{V}, v_y^\mathrm{V}, \dot{\psi}^\mathrm{V}, \delta_{\wheel})$ of the prediction model
and outputs optimal values of $\delta_{\wheel}^*$ and $\tau_{\wheel}^*$ to control the plant.
Here, $\tau_{\wheel}$ is the wheel torque and $\delta_{\wheel}, \dot{\delta}_{\wheel}$ are the steering angle and rate at
each wheel $\wheel\in\{\mathrm{FL}, \mathrm{FR}, \mathrm{RL}, \mathrm{RR}\}$.
$s$ is the Frenet arc length, $d$ is the lateral deviation, $\psi$ is the yaw angle,
and $v_x^\mathrm{V}, v_y^\mathrm{V}$ are the longitudinal and lateral velocities in the vehicle frame, respectively.
In \emph{IPG CarMaker}, the dynamics of the controlled vehicle given
$\tau_{\wheel}^*$ and $\delta_{\wheel}^*$ are simulated,
the resulting states are recorded and fed back into the control loop.
For example, a resulting trajectory of the simulated vehicle is shown in \autoref{fig:trajectory} (blue).
While not considered in this work, adding noise, delays, and other disturbances could be done in future work.
The deviation with respect to the reference trajectory can be measured through
\begin{align}
    \varepsilon_{\mathrm{long}} = |s - s{\refe}|,\,\, \varepsilon_{\mathrm{lat}} = |d|,\,\, \varepsilon_{\mathrm{yaw}} = |\psi - \psi{\refe}|
\end{align}
for the deviation in the longitudinal, lateral, and yaw direction, respectively.
All variables above depend on time $t$.
By computing $\varepsilon_{(\cdot), \max} := \max_t \varepsilon_{(\cdot)}(t)$,
we can check whether the controller violates a specific deviation limit.

\subsection{Definition of Degradation and Failure Parameters} \label{sec:degradations}
We particularly focus on the impact of degradations and failures of the vehicle's actuators on the controller's performance.
To cluster different degradation and failures, \textcite{stolte_toward_2023} introduce several failure and degradation categories.
They concern actuators that are relevant for motion control, i.e., steering, braking, and powertrain (and tire damage which is not in the focus of the present paper).
In a vehicle dynamics model, a degradation or failure can be expressed in the easiest case
through a limit or enforced value setting (e.g., to zero) of a specific model variable.
In this work, we focus on degradations and failures of \mobile's steering, powertrain and braking systems, %
and hence, we represent a degradation or failure as a limited value (range)
of the steering angle $\delta_{\wheel}$, steering rate $\dot{\delta}_{\wheel}$, and/or wheel torque $\tau_{\wheel}$,
which lays within (or is a subset of) the nominal value range.
For the sake of brevity, we use $z\in \{\delta, \dot{\delta}, \tau\}$.
We then express such a degradation or failure of an actuator as
\begin{equation}
    z_{\wheel}^{\min} \leq z_{\degradation, \wheel}^{\min} \leq z_{\wheel}\leq z_{\degradation, \wheel}^{\max} \leq z_{\wheel}^{\max},
\end{equation}
with %
$z_{\wheel}^{\min}$ and $z_{\wheel}^{\max}$
indicating the minimum and maximum values of a variable due to (nominal) technical limitations,
and $z_{\degradation, \wheel}^{\min}$ and $z_{\degradation, \wheel}^{\max}$ indicating degraded value ranges.

We assume that $z_{\wheel}$ can take arbitrary values within the specified range.
A \qemph{failure} is hence represented, e.g., through one parameter pair being set to zero.
In the following, we only consider this type of failure,
although other fixed values would also fulfill the criterion of a loss of functionality.
In the following, we do not further distinguish between degradations and failures,
and only use the term \qemph{degradation parameter}.

Hereafter, we consider $|z|_{\degradation,\wheel} := \max\{|z_{\degradation, \wheel}^{\min}|, |z_{\degradation, \wheel}^{\max}|\}$.
This is an overestimation of the effects of degradations that would only affect the positive or negative direction.
Then, we can define the normalized degradation parameter
\begin{equation}
    \zdegwheel:=\frac{|z|_{\degradation,\wheel}}{z_{\wheel}^{\max}} \in [0, 1],
\end{equation}
that is normalized by its nominal (i.e., due to technical limitations) limit $z_{\wheel}^{\max}$,
with $\delta_{\wheel}^{\max}=30\deg$, $\dot{\delta}_{\wheel}^{\max}=120\degs$ and $\tau_{\wheel}^{\max}=2000\Nm$,
see \parencite{stolte_toward_2023}.
\section{Conformalized Prediction Model} \label{sec:predictor}
Using the data gained during the simulation runs, we develop a data-based model
that forecasts the controller's deviation during a specific maneuver under current conditions
-- including potential actuator degradations and failures.
At runtime, we aim to identify those actions that are likely to result in a sufficiently low deviation
by exploring a finite set of possible maneuver parameters,
given the availability of environmental and FDI information.
Hereafter, we consider a lane change maneuver and focus on a single deviation metric as an example.
We aim to identify
\begin{equation}
    y = f(\wmax, \wmin, \kmax, \, \dots,\, \deltadotdegwheel,\, \taudegwheel) =: f(\inputnn),
\end{equation}
with the input vector $\inputnn$ and the lateral deviation $y = \epsNmax$ as output,
as described in \autoref{tab:mlp_inputs_outputs}.
We neglect $r_{\odd}$ and $\mu_{\odd}$ as we limit our experiments to
$r_{\odd} = \mathrm{left}$ and $\mu_{\odd} = 1$ in the following.
In this paper, we choose a \emph{neural network} to approximate $f$ with $\hat{y} = \hat{f}(\inputnn)\approx f(\inputnn)$.
If considering multiple metrics would be desired,
either training multiple models or using a multi-output model is possible.
Alternatively, similar to \parencite{mcconachie_learning_2020}, a classification approach could be adopted,
i.e., using a binary state indicating whether a deviation threshold is exceeded.
However, the regression model is more flexible as classification could be added as an additional layer.
\begin{table}
    \centering
    \vspace{0.35em}
    \caption{Neural Network Architecture and Training Parameters}
    \renewcommand{\arraystretch}{1.2}
    \begin{adjustbox}{width=0.94\linewidth}
        \begin{minipage}[t]{0.55\linewidth}
            \centering
            \begin{tabular}{ccc}
                \toprule
                \textbf{Layer} & \textbf{Neurons} & \textbf{Activation} \\
                \midrule
                Linear 1 & 19 & ReLU \\
                \rule{0pt}{4.08ex} %
                Linear 2 & 209 & ReLU \\
                \rule{0pt}{4.08ex} %
                Linear 3 & 209 & ReLU \\
                \rule{0pt}{4.08ex} %
                Linear 4 & 2 & ReLU \\
                \bottomrule
            \end{tabular}
        \end{minipage}%
        \begin{minipage}[t]{0.025\linewidth}
            \centering
            \vspace{0pt}
        \end{minipage}%
        \begin{minipage}[t]{0.55\linewidth}
            \centering
            \begin{tabular}{cc}
                \toprule
                \textbf{Parameter} & \textbf{Value} \\
                \midrule
                Batch Size & 64 \\
                Optimizer & RMSprop \\
                Learning Rate & 0.0005 \\
                Num. of Epochs & 1200 \\
                Early Stopping & True \\
                Normalization & Batch Norm \\
                \bottomrule
            \end{tabular}
        \end{minipage}
    \end{adjustbox}
    \label{tab:mlp_arch}
\end{table}

Using a deterministic model suggests that any prediction could be made with certainty.
However, several sources of uncertainty exist:
For example, we only use selected indicators to express the road segment's geometry,
e.g., min./max. curvature and width. %
Additionally, we find that the controller's numerical optimization procedure
sometimes tends to find worse solutions for slightly less restricted problems than for more restricted ones,
i.e., the deviation sometimes declines when vehicle dynamics demand and/or degradation or failure severity rise.
Especially in dynamically challenging scenarios in the presence of degradations and failures,
the controller has its difficulties to keep the vehicle on the desired trajectory,
resulting in very large and fluctuating (across trajectories) deviation values.
In such cases, it is also possible that even the commercial-grade simulation model reaches its limits.
Therefore, we find a \emph{stochastic} approach to be more suitable for our problem.
Also, other sources of uncertainty
-- such as measurement and actuator noise -- %
should be covered by the selected approach if investigated in future work.

\subsection{Conformalized Quantile Regression} \label{sec:cqr}
We adopt a \qemph{quantile regression} \parencite{koenker_regression_1978}
approach, which, instead of predicting a single value $\hat{y}\approx y$,
results in an interval $\interval(\inputnntest) := [\hat{y}_{\sulow},\, \hat{y}_{\suup}]$
for a test sample $\inputnntest\in\inputnntestset$ in which the true target value is expected to lie with a certain probability.
The distance between $\hat{y}_{\sulow}$ and $\hat{y}_{\suup}$ is referred to as the \qemph{interval length} $|\interval|$.
As ground truth, we consider the results of our simulation.
In particular, we use the \qemph{conformalized quantile regression} (CQR) method
\parencite{romano_conformalized_2019, angelopoulos_gentle_2022}, which is a method
for constructing prediction intervals for a given significance level $\alpha$
that are guaranteed to have the correct \qemph{coverage (probability)} $\coverage$,
\begin{equation} \label{eq:conformal_prediction}
    \prob\left \{\outputnntest \in \interval(\inputnntest) = [\hat{y}_{\sulow},\, \hat{y}_{\suup}] \right \} \geq 1-\alpha =: \coverage.
\end{equation}
The calibration of the prediction intervals is achieved using a calibration dataset $\inputnncalset$
and holds on unseen test data $\inputnntestset$ if it is independent and identically distributed with respect to $\inputnncalset$.
For uncertainty estimation of neural network predictions,
CQR is particularly appealing as it is distribution-free, %
adapts well to heteroscedastic data \parencite{romano_conformalized_2019}
and yields the aforementioned guarantee.
Of course, this guarantee is only useful if the data used for calibration and evaluation
is representative with respect to the system's real dynamics --
which can only be achieved by using real-world data.

\subsection{Dataset Creation} \label{sec:dataset}
\begin{figure}[H]
    \vspace{-1.0em}
    \centering
    \includegraphics[width=0.925\linewidth]{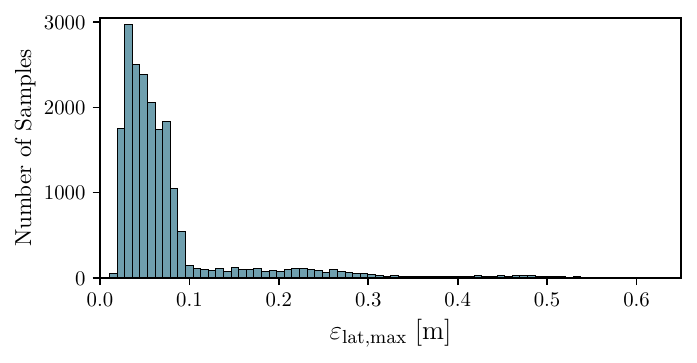}
    \vspace{-1.25em}
    \caption{Distribution of $\epsNmax$ in the dataset}
    \label{fig:distribution}
    \vspace{-1.0em}
\end{figure}
To generate training and evaluation datasets, we use previous simulation results.
Given the set of real road segments described in \autoref{sec:refgen},
we randomly pick $\nsegments$ segments and generate lane change trajectories on each of them.
Next, we employ the offline reference trajectory generator
and randomly sample maneuver parameters from the value ranges in \autoref{tab:mlp_inputs_outputs}.
In terms of the maneuver parameters, it would be feasible to use data from real-world
test drives within the desired ODD to generate a representative distribution.
However, we note that dynamically challenging conditions are rarely encountered in real-world scenarios
and therefore choose to draw the maneuver parameters from a uniform distribution,
limited by the value ranges in \autoref{tab:mlp_inputs_outputs}.
Any conclusions drawn from this dataset are hence rather conservative.

We receive $\nmaneuver$ different sets of maneuver parameters and
generate $\nsegments\cdot \nmaneuver$ target trajectories in total.
Finally, we additionally generate random sets of degradations and failures
and simulate the trajectory tracking behavior for the nominal case
as well as for each degradation and failure.
We aim to design a predictor that works well
even in rarely occurring degradation events and choose uniform distributions accordingly.
We generate $\ndeg$ different sets of degradation parameters, including the nominal case.
In the remainder of this paper, we set $\nsegments=50$, $\nmaneuver=20$, $\ndeg=20$,
and receive $\nsegments\cdot \nmaneuver\cdot \ndeg = 20000$ samples. %
For each sample, $\epsNmax$ is measured and stored. %
An extract of the distribution is shown in \autoref{fig:distribution}.
The frequency of large deviations is low due to
the fault tolerance of the controller, the over-actuation of the vehicle,
and the fact that we include the nominal case for each reference trajectory.
In line with \parencite{da_silva_crash-prone_2024},
we find that larger values of $\epsNmax$ are more likely to occur
when the maneuver is challenging and/or actuators are degraded.

\subsection{Model Training and Evaluation} \label{sec:training_evaluation}
The neural network and CQR procedure are implemented using the \pytorch\ library,
relying on the implementation of \textcite{romano_conformalized_2019}.\footnote{%
    {\smallerfootnote\url{https://github.com/yromano/cqr}}, accessed 2024-03-25.
}
The data is split into a training and a testing set during training with a ratio of $80\%$ to $20\%$,
i.e., $|\inputnntrainset| = 16000$ and $|\inputnntestset| = 4000$.
Of the latter, $|\inputnncalset| = 2000$ samples are held out for calibration.
Standard scaling is applied to all inputs.
While the guarantee in \autoref{eq:conformal_prediction} holds
for any prediction model to which the calibration in \parencite{romano_conformalized_2019} is applied,
better models will result in shorter prediction intervals \parencite{angelopoulos_gentle_2022}.
We therefore conduct a grid search to find optimal hyperparameters for the neural network,
i.e., those that minimize the used ``pinball loss'' \parencite{koenker_regression_1978, romano_conformalized_2019}.
For $\coverage = 99\%$, the network architecture that minimizes training loss is shown in \autoref{tab:mlp_arch}.
For different values of $\alpha$, we use the same network architecture but \mbox{re-train} the model.
\begin{table}[H]
    \vspace{-0.5em}
    \centering
    \caption{Empirical Coverage and Interval Lengths on $\inputnntestset$}
    \begin{adjustbox}{center, keepaspectratio, width=0.85\columnwidth}
    \renewcommand{\arraystretch}{1.2}
    \begin{tabular}{p{1.7cm}ccccccc}
        \toprule
        Desired $\mathbf{\coverage}$ & $[\mathrm{\%}]$ & $\mathbf{90.0}$ & $\mathbf{92.5}$ & $\mathbf{95.0}$ & $\mathbf{97.5}$ & $\mathbf{99.0}$ & $\mathbf{99.5}$ \\
        \midrule
        Empirical $\empcoverage$ & $[\mathrm{\%}]$      & $88.40$ & $91.20$ & $93.72$ & $96.67$ & $99.14$ & $99.52$ \\
        $\quantile_{10\%}(|\interval|)$ & $[\mathrm{cm}]$ & $2.30$ & $1.95$ & $2.91$ & $1.73$ & $5.09$ & $6.01$ \\
        $\median(|\interval|)$ & $[\mathrm{cm}]$          & $5.20$ & $4.76$ & $5.73$ & $4.84$ & $8.21$ & $9.56$ \\
        $\mean(|\interval|)$ & $[\mathrm{cm}]$            & $9.90$ & $8.25$ & $10.24$ & $10.68$ & $15.98$ & $19.32$ \\
        $\quantile_{90\%}(|\interval|)$ & $[\mathrm{cm}]$ & $25.25$ & $19.20$ & $25.23$ & $31.50$ & $41.98$ & $53.49$ \\
        $\quantile_{10\%}(\Delta\interval_{\mathrm{Hi}})$ & $[\mathrm{cm}]$ & $0.15$ & $0.32$ & $0.37$ & $0.54$ & $1.54$ & $3.43$ \\
        $\median(\Delta\interval_{\mathrm{Hi}})$ & $[\mathrm{cm}]$          & $0.58$ & $0.72$ & $0.82$ & $1.16$ & $2.85$ & $5.09$ \\
        $\mean(\Delta\interval_{\mathrm{Hi}})$ & $[\mathrm{cm}]$            & $2.93$ & $1.86$ & $3.21$ & $4.51$ & $7.80$ & $12.16$ \\
        $\quantile_{90\%}(\Delta\interval_{\mathrm{Hi}})$ & $[\mathrm{cm}]$ & $8.12$ & $4.10$ & $6.76$ & $13.21$ & $22.77$ & $35.21$ \\
        \bottomrule
    \end{tabular}
    \end{adjustbox}
    \label{tab:mlp_results}
    \vspace{-1.0em}
\end{table}

The results on the test set $\inputnntestset$ are given in \autoref{tab:mlp_results}.
We see that the empirically observed coverage $\empcoverage$ is close to the desired coverage $\coverage$
(with a maximum deviation of $1.60\%$) across all desired values of $\coverage$.
As expected, the mean of the interval length $|\interval|$
as well as its median grow when increasing the value of $\coverage$.
With respect to the distribution shown in \autoref{fig:distribution},
we can conclude that the intervals are very tight
for large portions of the data, i.e., for small values of $\epsNmax$.
Given in-depth analyses, we find that larger intervals appear for high values of $\epsNmax$.
For later applications, we will focus on $\epsNmaxHi$ as it is the upper limit of the interval
in which we assume $\epsNmax$ to lie with a probability of $\leq\alpha$.
To evaluate the degree of overestimation of the CQR method,
we compute the difference $\Delta I_{\mathrm{Hi}} := \epsNmaxHi - \epsNmax$
for those cases where $I$ actually covers the true value of $\epsNmax$.
As for $|\interval|$, we find that the values of $\Delta I_{\mathrm{Hi}}$
increase for an increasing value of $\coverage$.
We once again find that for higher true values of $\epsNmax$,
the predictor tends to overestimate. %
\section{Application of the Prediction Model} \label{sec:application}
\begin{figure}
    \vspace{0.35em}
    \centering
    \includegraphics[width=\columnwidth]{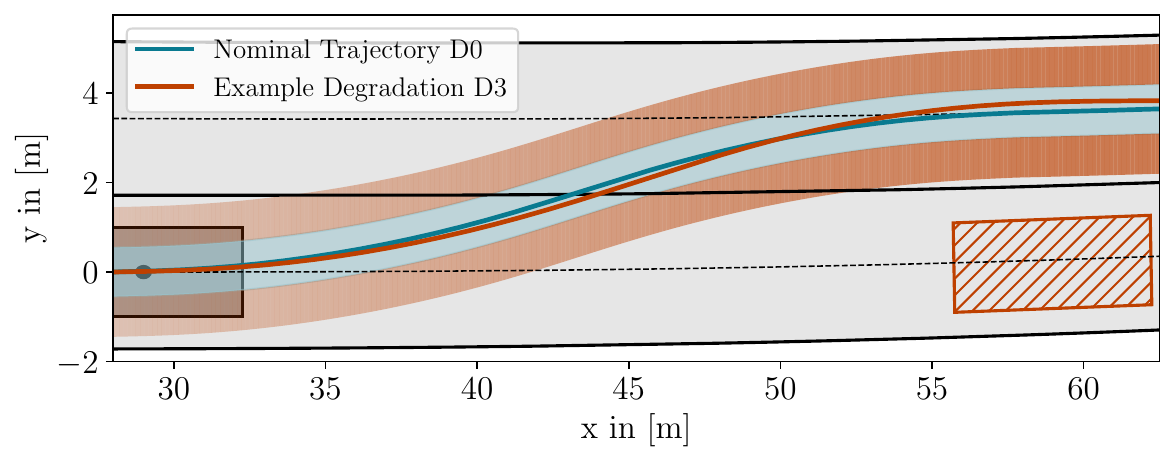}
    \vspace{-2.0em}
    \caption{%
        Nominal path (blue) for a lane change maneuver at $\amax = 3\tmss$
        compared to the resulting path (red) in case of %
        $\mathrm{D}3$ (see \autoref{tab:deg_examples}).
        Blue area indicates lateral spaced covered by prediction $\pm\epsNmaxHi$ with $\coveragetwo = 99\%$.
        Light red area indicates the additional space covered by the ego-vehicle $\pm\wveh/2$.
        Red shaded area indicates occupancy by an obstacle.%
        }
    \label{fig:trajectory}
\end{figure}
To illustrate the application of the trained prediction model,
we consider an example scenario in which the automated vehicle
enters a new road segment with two lanes and
unexpectedly encounters an obstacle within the lane, i.e., a stopped vehicle.
The automated vehicle may choose between two evasive maneuvers:
(a) changing lanes to the left or (b) to stop in front of the obstacle.
Assuming that the vehicle is able to determine its current dynamic
state, degradation and failure information, and relevant operating conditions
-- such as the current road segment's geometry --
the vehicle shall make an informed decision.
\begin{table}[H]
    \vspace{-0.5em}
    \centering
    \caption{Example Degradations $\mathrm{D}0\dots\mathrm{D}5$ as Random Fixed Settings
        of the Degradation Parameters $\zdegwheel$ for the Example Scenario. %
        Blue color indicates higher and red color lower values.}
    \vspace{-1.0em}
    \begin{adjustbox}{width=0.97\columnwidth, center}
        \includegraphics[width=1.0\columnwidth]{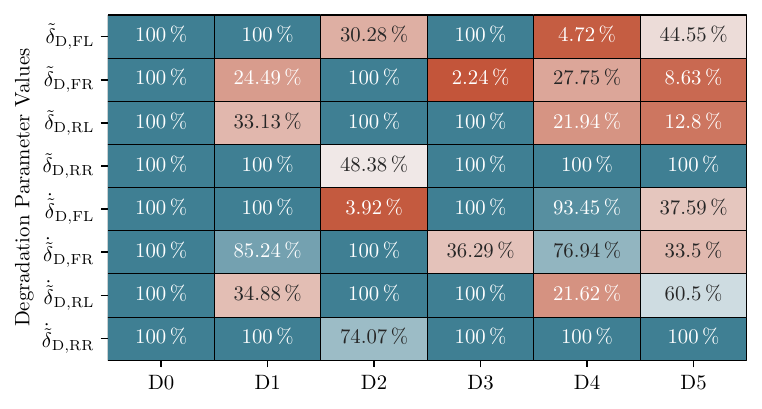}
    \end{adjustbox}
    \label{tab:deg_examples}
    \vspace{-2.25em}
\end{table}
We assume the vehicle's current speed to be $\vinitial = 50\tkmh$,
which meets the German inner-city speed limit.
While many parameters are set due to the initial scene
(such as geometrical characteristics of the road segment
and the vehicle's degradation and failure state),
the vehicle may still choose different maneuver options,
apply different acceleration limits and target speeds.
For the sake of brevity, we only consider degradations of the steering actuators
and their impact on the vehicle's lateral dynamics during
a possible lane change maneuver at different acceleration limits.
We further assume $\vtarget = 50\tkmh$ and $\mu_{\odd} = 1$.

In order to inform the behavior generation, we make the following considerations:
We consider the current value of $\wmin = 2.78\m$ for the road segment's minimum width
and the width $\wveh = 1.96\m$ of our test vehicle \mobile .
Given the worst case assumption that the maximum lateral deviation is reached where the road segment is narrowest,
the free lateral space is $\epsNmaxFree = 41\tcm$ both on the left and right side.
Accordingly, the vehicle should not execute maneuvers during
which a lateral deviation of $\epsNmaxFree$ is exceeded.
Hereafter, we only consider the upper prediction bound $\epsNmaxHi$ as described.
We visualize our approach in \autoref{fig:trajectory} --
where it becomes clear that the application of CQR leads to an expected ``corridor'' of lateral control deviation.
However, in addition to the controller's performance,
it is crucial to note that the planned trajectory
may not ideally connect the current and target lane center,
e.g., due to additional uncertainty in the environmental perception
and/or map-based information.
Furthermore, disturbances in the vehicle's state estimation may arise.
Both will require additional lateral margins %
that we do not address in our framework.
\begin{table}[H]
    \centering
    \vspace{-0.5em}
    \caption{Prediction $\epsNmaxHi$ vs. true lateral deviation $\epsNmax$ in $[\mathrm{m}]$. %
        Blue color indicates low and red color indicates high true values. %
        For $\coverageone = 95\,\%$ (top) and $\coveragetwo = 99\,\%$ (bottom).}
    \vspace{-1.0em}
    \begin{adjustbox}{width=0.97\columnwidth, center}
        \includegraphics[width=1.0\columnwidth]{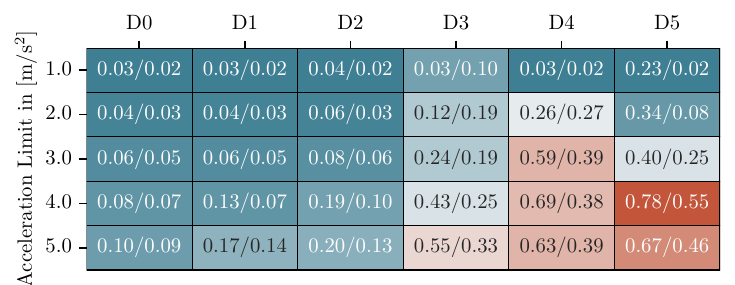}
    \end{adjustbox}
    \label{tab:pivot_1}
    \vspace{-1.0em}
    \begin{adjustbox}{width=0.99\columnwidth, center}
        \includegraphics[width=1.0\columnwidth]{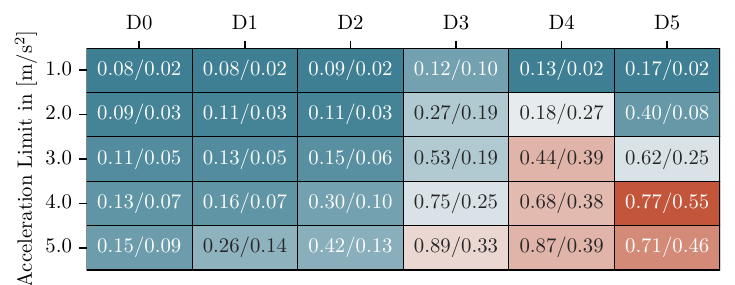}
    \end{adjustbox}
    \label{tab:pivot_2}
    \vspace{-2.0em}
\end{table}
At runtime, the automated vehicle can use the prediction model and is able
to feed possible maneuver (parameter) options into the model,
in order to predict the expected bounds of the lateral deviation.
Also, we assess the impact of different degradations that are defined in \autoref{tab:deg_examples}.
In \autoref{tab:pivot_1}, we present the results of the prediction model:
The predicted values of $\epsNmaxHi$ are visualized along with ground truth values of $\epsNmax$ from the simulation.
As an example, we set $\coverageone = 95\%$ (top) and $\coveragetwo = 99\%$ (bottom).
In \autoref{tab:pivot_1}, given the severity of the respective degradation $\mathrm{D}0\dots\mathrm{D}5$
as well as the acceleration limit $\amax$, %
we see strong changes in both the actual and predicted deviation.
For some cases -- even for high accelerations -- the expected deviations is much smaller than the value of $\epsNmaxFree$,
resulting in a remaining margin of $\epsNmaxFree - \epsNmaxHi$,
which might be required to take account for additional uncertainty in the closed loop.
For $\mathrm{D}4,\,\mathrm{D}5$, we see great deviations from the reference trajectory in terms of $\epsNmax$.

Especially for high accelerations, the vehicle's lateral dynamics are significantly impacted by the degradation.
Furthermore, the predicted values of $\epsNmaxHi$ are very high in these cases and greatly exceeds the true value of $\epsNmax$. %
This is necessary to take account of the uncertainty as described in previous sections.
Given \autoref{tab:mlp_results}, it becomes clear that the degree of overestimation is related to the value of $\alpha$.
To illustrate that, we consider the example of performing a lane change with $\amax = 3\tmss$ given degradation $\mathrm{D}3$:
The model trained for $\coverageone = 95\,\%$ %
predicts a maximum lateral deviation of up to $0.24\m$, %
thus exceeding the true value of $\epsNmax = 0.19\tcm$.
However, the prediction $\epsNmaxHi$ is still smaller than $\epsNmaxFree$ and hence,
the vehicle's action space is not necessarily limited due to an overly conservative prediction
(however, it might be when considering additional uncertainty in other system parts).
In terms of $\coveragetwo$ %
we see that the predicted value is even higher,
$\epsNmaxHi = 0.53\m$, and therefore larger than $\epsNmaxFree$, i.e.,
the corresponding lane change should be rated as infeasible given the prediction.
Finetuning the value for $\alpha$ is therefore crucial for the behavior of the prediction model. %
\section{Conclusion and Future Work} \label{sec:conclusion}
In this work, we aim to create a self-representation model
of an automated vehicle's motion controller in a data-driven manner.
Using an offline simulation, we generate a set of trajectories
and train a neural network to predict the maximum lateral deviation introduced by the controller
given the maneuver to be executed, operational conditions, and potential actuator degradations or failures.
Using conformalized quantile regression, we acknowledge the uncertainty in the prediction
and provide prediction intervals for the deviation metric that meet a given significance level.
We demonstrate that our predictor is able to estimate the controller's performance
with the expected coverage and with low interval length for most cases.
Our results show that the feasibility of executing a specific lateral maneuver is
a function of the vehicle's operational conditions
(i.e., the available free lateral space), its dynamic state, and the presence of a actuator degradation or failure.
Our method allows to assess this intuitive relation quantitatively and with the required statistical rigor.
Overall, while empirical and desired coverages align well,
we find that our method is rather conservative.
While conservatism is often desired in safety-critical applications,
future work should focus on reducing the average interval length
in order to not unnecessarily restrict the vehicle's action space.

While focusing on a lane change maneuver and the vehicle's lateral deviation as an example,
we state that with a more complex predictor that covers multiple maneuvers
as well as additional performance/deviation metrics,
the behavior generation process can be informed more comprehensively.
However, we cannot rule out the possibility
that when applying threshold values for different performance measures,
the controller's performance could potentially be insufficient for any maneuver option under certain conditions.
The behavior generation in such cases can be informed by the predictor but not resolved.

Furthermore, we once again emphasize that further considerations are required to set an appropriate value for $\coverage$.
While the statistical representation allows us to inform a stochastic decision-making framework,
defining such a value requires a decision with respect to the acceptable level of uncertainty, which is out of the scope of this paper.
While the prediction intervals exemplified in this work maintain a feasible length, %
large values of $\coverage$ lead to longer prediction intervals which might render the predictor useless in most scenarios,
requiring an enhanced design thereof, e.g., involving more input parameters
or a sequential model instead of a sparse ``one-shot'' feed-forward model.
The same might apply to the impact of additional disturbances.

Finally, conducting our analysis and model training using real-world data
is a crucial step to ensure that the predictor's output is reliable in real-world experiments,
e.g., using a closed test track and high fidelity state estimation to safely acquire ground truth data.
Besides the need for more realistic data, we also look forward to integrating
several sources of uncertainty into our simulation, leading to new challenges. %
\section*{Acknowledgement} \label{sec:acknowledgement}
We thank Jens Rieken for providing the road segment data.
We thank Xiao Yun for his support with the implementation.
\printbibliography
\end{document}